\documentclass[runningheads]{llncs}
\usepackage[T1]{fontenc}
\usepackage{cite}
\usepackage{graphicx}
\usepackage{amsmath,amssymb,amsfonts}
\usepackage{hhline}

\DeclareMathOperator*{\argmax}{arg\,max}

\begin{document}
\title{End-to-End Training of Back-Translation Framework with Categorical Reparameterization Trick}
\titlerunning{E2E Training of Back-Translation with CRT}
%
\author{DongNyeong Heo\inst{1}\orcidID{0000-0002-8765-6744} \and
Heeyoul Choi\inst{1}\orcidID{0000-0002-0855-8725}}
%
\authorrunning{D. Heo and H. Choi}
%
\institute{Department of Computer Science and Electrical Engineering, \\
Handong Global University, Pohang, South Korea \\
\email{sjglsks@gmail.com, hchoi@handong.edu}
}
\maketitle              
\begin{abstract}
Back-translation (BT) is an effective semi-supervised learning framework in neural machine translation (NMT). A pre-trained NMT model translates monolingual sentences and makes synthetic bilingual sentence pairs for the training of the other NMT model, and vice versa. Understanding the two NMT models as inference and generation models, respectively, the training method of variational auto-encoder (VAE) was applied in previous works, which is a mainstream framework of generative models. However, the discrete property of translated sentences prevents gradient information from flowing between the two NMT models. In this paper, we propose the {\it categorical reparameterization trick (CRT)} that makes NMT models generate {\it differentiable sentences} so that the VAE's training framework can work in an end-to-end fashion. Our BT experiment conducted on a WMT benchmark dataset demonstrates the superiority of our proposed CRT compared to the Gumbel-softmax trick, which is a popular reparameterization method for categorical variable. 
Moreover, our experiments conducted on multiple WMT benchmark datasets demonstrate that our proposed end-to-end training framework is effective in terms of BLEU scores not only compared to its counterpart baseline which is not trained in an end-to-end fashion, but also compared to other previous BT works. The code is available at the web\footnote{https://github.com/Nunpuking/End-To-End-Backtranslation}.

\end{abstract}
\section{Introduction}
\label{sec:intro}
Supervised learning algorithms in the neural machine translation (NMT) task have shown outstanding performances along with successes in deep learning \cite{bahdanau2014neural, vaswani2017attention}. Those algorithms perform well if there is a large amount of bilingual corpus. However, just a few pairs of languages have large bilingual corpora, while most of the other pairs do not. In addition, even though a language pair has a large bilingual corpus, it should be updated on a regular basis because language is not static over time. New words appear, and existing words might disappear, corresponding to changes in culture, society, and generations. Therefore, supervised learning algorithms for the NMT task suffer endlessly from a data-hungry situation and expensive data collection for bilingual corpus. 

Unlike bilingual corpus, a monolingual corpus is easy to collect. Therefore, semi-supervised learning algorithms that use additional monolingual corpora in various ways have been suggested \cite{gulcehre2015using, zhang2016exploiting, currey2017copied, domhan2017using, skorokhodov2018semi}. Alongside these methods, the back-translation (BT) methods have been proposed with showing significant performance improvements from the supervised learning algorithms \cite{sennrich2015improving, he2016dual, hoang2018iterative, edunov2018understanding, zhang2018joint, xu2020dual, guo2021revisiting}.

The central idea of the BT method is firstly proposed by \cite{sennrich2015improving}. The pre-trained target-to-source NMT (TS-NMT) model, which is trained with only a bilingual corpus, translates target-language monolingual sentences to source-language sentences. Then, the translated source-language sentences are used as the synthetic pairs of the corresponding target-language monolingual input sentences. By adding these synthetic pairs to the original bilingual corpus, the size of the total training corpus increases, like the data augmentation \cite{Goodfellow-et-al-2016}. Then, this increased corpus is used to train the source-to-target NMT (ST-NMT) model. The same process with source-language monolingual sentences can be applied to train the TS-NMT model.

The theoretical background of the BT method has been developed based on the auto-encoding framework \cite{cotterell2018explaining, zhang2018joint, xu2020dual}, such as variational auto-encoder (VAE) \cite{kingma2013auto}. Considering the translated sentence as the inferred latent variable of the corresponding monolingual input sentence, BT can be understood as a reconstruction process. For example, the TS-NMT model infers a \textit{`latent sentence'} in the source-language domain given a target-language monolingual sentence as an input, then the ST-NMT model reconstructs the input target-language monolingual sentence from the latent sentence. This process is a target-to-source-to-target (TST) process. In this process, the TS-NMT model approximates the posterior of the latent sentence as an \textit{`inference model'}, and the ST-NMT model estimates the likelihood of the monolingual sentence as a \textit{`generation model'}. Likewise, the source-to-target-to-source (STS) process is conducted with the opposite order and roles with source-language monolingual corpus.

However, training the NMT models in the VAE framework is challenging for several reasons. First, the distribution of each word in the latent sentence is discrete categorical distribution, and the non-differentiable latent sentence makes the backpropagation impossible to train the inference model. Second, the latent sentence should be a realistic sentence in that language domain, not an arbitrary sentence, to guarantee translation quality. Note that in the conventional VAE models, the latent space is modeled to have an isotropic Gaussian distribution without any other regularizations on the space, though there are several works that regularize the space to disentangle the dimensions \cite{Higgins2017betaVAELB, kim2018disentangling, hahn2019disentangling}. If we do not address this issue, the inference model can be trained to mistranslate the monolingual input sentence, just focusing on trivial reconstruction. Because of these challenges, previous works trained only the generation model or used expectation-maximization (EM) \cite{cotterell2018explaining,zhang2018joint} algorithm. However, such optimization methods are not effective because the inference model cannot update its parameters directly along with the generation model with respect to the final objective function.

\begin{figure}[t!]
    \centering
    \includegraphics[width=0.6\linewidth]{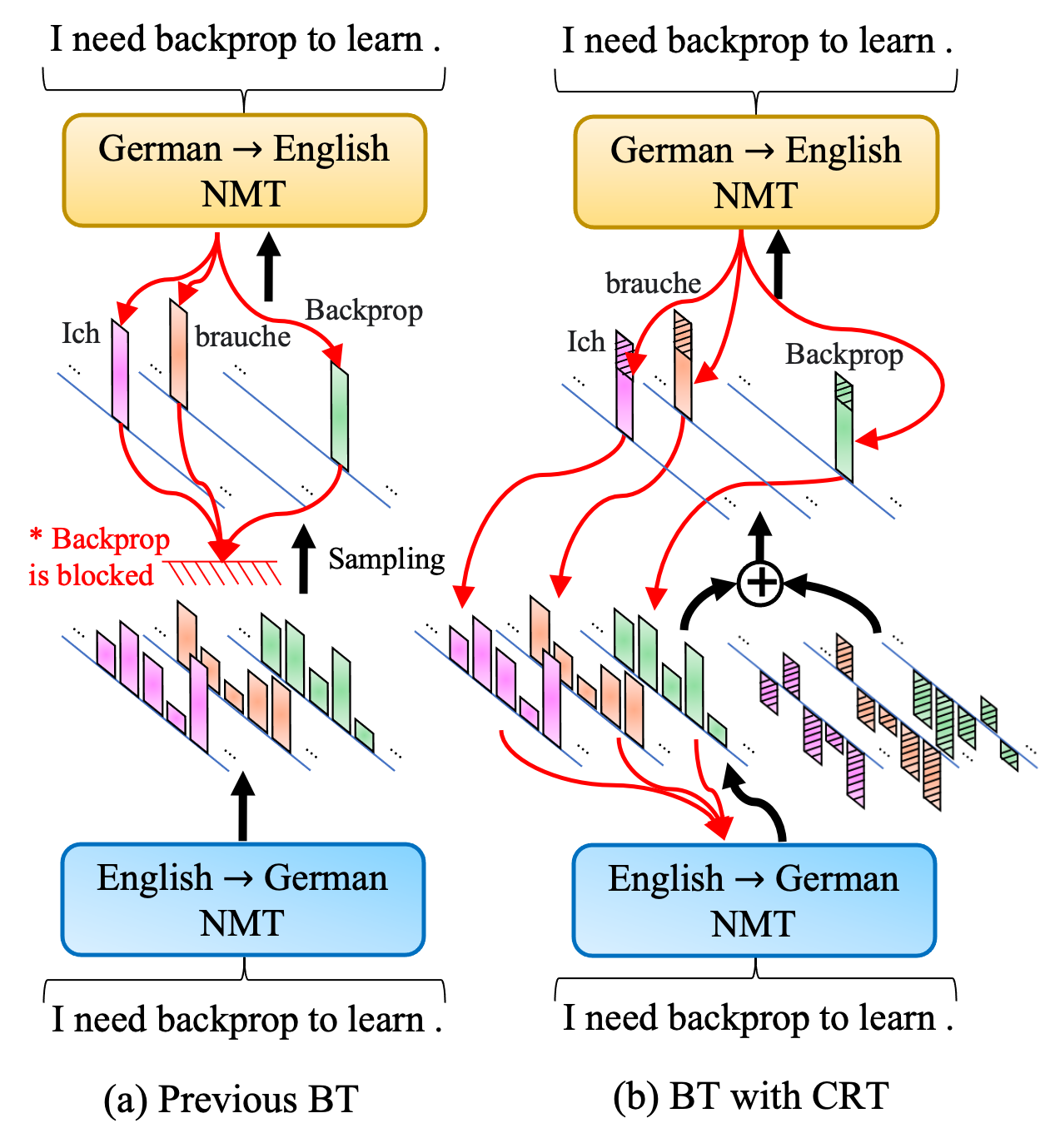}
    \caption{Examples of \textit{English-to-German-to-English} monolingual process of (a) previous BT and (b) BT with our proposed CRT. Given the English sentence, \textit{'I need backprop to learn.'}, they infer the German sentence, \textit{'Ich brauche Backprop, um zu lemen.'} (only the first three words are illustrated for brevity). The solid black lines are forward propagations, and the thin red lines are backward propagations. While the sampling operation blocks the gradient flow in the previous works (a), the gradient flow detours to the inference model (English-to-German) in (b).
    }
    \label{fig:illustration_our_method}
\end{figure}

In this paper, we propose a new algorithm that handles the above challenges so that the two NMT models in the BT framework can be trained by end-to-end backpropagation, as in VAE. To overcome the non-differentiable issue, we propose a categorical reparameterization trick (CRT). By adding a conjugate part to the estimated categorical distribution, the CRT outputs an \textit{differentiable sentence}, skipping the non-differentiable sampling process. Therefore, the end-to-end backpropagation finally updates the inference model to directly minimize with respect to the final objective function. The comparison of the previous BT works and the BT with our CRT is illustrated in Fig.~\ref{fig:illustration_our_method}. Additionally, we study the advantages of the CRT compared to the Gumbel-softmax trick (GST) \cite{jang2017categorical}, which is a popular reparameterization method for categorical distribution. In addition, in order to regularize the latent sentence to be realistic, we use the output distribution of the pre-trained language model \cite{bengio2003neural,Paszke2019pytorch} as the prior distribution of the latent sentence. Finally, we propose a regularization technique that controls the amount of stochasticity in the inference process of latent sentence during training.


Our experiments conducted on multiple NMT benchmark datasets, such as WMT14 English-German and WMT18 English-Turkish \cite{rej2018findings}, demonstrate the advantages of our proposed approaches compared to the baseline BT method, which was not trained using end-to-end backpropagation. Moreover, we compare our proposed approach with previous works that used comparable training setting scales (e.g., model size and dataset size). The results show considerable performance gains of our approach compared to previous BT methods.

\section{Related Works}
\subsection{Back-translation (BT) for NMT}
\label{subsec:related_works_monolingual_NMT}
\subsubsection{Original BT}
Using a monolingual corpus for the NMT task often improves translation performance because it can give additional information for language understanding. As a practical method, the BT method was proposed \cite{sennrich2015improving, imamura2018enhancement, edunov2018understanding, fadaee2018back, gracca2019generalizing, caswell2019tagged} and the main idea is augmenting the bilingual corpus with synthetic pairs. During the training of ST-NMT, a pre-trained TS-NMT model which was pre-trained by only bilingual corpus, translates target-language monolingual sentences into source-language synthetic sentences. Then, the synthetic source-language sentences and the corresponding target-language monolingual input sentences become synthetic pairs, then used as additional bilingual sentence pairs.



\subsubsection{Iterative BT}
Motivated by the above approach, the iterative back-translation (IBT) method that symmetrically conducts the original BT method multiple times was proposed \cite{hoang2018iterative}. In other words, given both pre-trained models of ST-NMT and TS-NMT, they translate and make synthetic pairs of source-language and target-language monolingual corpora, respectively. Each synthetic pair that made by a NMT model is used to train the other NMT model. This process iteratively operates, so that the quality of the synthetic pairs can be improved by the improved NMT models. Repetition of this process often significantly outperform the original BT method. The IBT method can be divided into three ways according to the update timings of synthetic sentences: offline IBT, online IBT, and semi-online IBT. Offline IBT updates the whole synthetic sentences after the models converge within a single BT process. Different from the offline IBT, online IBT updates a mini-batch of the synthetic sentences at every iteration. Semi-online IBT follows the same strategy as the online IBT, but it sometimes loads previous synthetic sentences from memory given a pre-defined probability \cite{han2021unsupervised}. In this paper, we follow the semi-online IBT.

\subsubsection{Probabilistic Framework of IBT}
Following the IBT method, a probabilistic framework was proposed considering the translated synthetic sentence as an inferred latent variable that is the aligned representation of the monolingual input sentence \cite{cotterell2018explaining, zhang2018joint}. As described in Section~\ref{sec:intro}, this BT framework is naturally connected to VAE based on the auto-encoding framework. For the sake of the reader's understanding, we repeatedly summarize important terminologies in this paper. In the case of the source-to-target-to-source (STS) process, the ST-NMT model plays the role of the \textit{inference model} (approximated posterior estimator), and it infers a \textit{latent sentence} in the target-language domain given a source-language monolingual sentence. The TS-NMT model plays the role of the \textit{generation model} (likelihood estimator), and it generates the input source-language monolingual sentence. In the target-to-source-to-target (TST) process, ST-NMT and TS-NMT play opposite roles.

However, it is still challenging to train the NMT models in the BT framework because of the non-differentiable latent sentences in the inference stage. Previous works proposed expectation-maximization algorithm \cite{cotterell2018explaining, zhang2018joint} or backpropagation with ignoring the update of the inference model \cite{xu2020dual}. In these cases, the inference model loses the learning signal that would have been propagated from the generation model. Therefore, it might find a worse local optimum because of the inefficient optimization. In this paper, we propose a new trick that reparameterizes the inferred latent sentence so that the end-to-end backpropagation can be feasible as in conventional VAE.

\subsection{Binary Reparameterization Trick}
\label{subsec:Binary_Reparam_Trick}
Learning discrete representations in neural networks has several advantages. First, it is proper to represent discrete variables such as characters or words in natural language. Second, it can be efficiently implemented at the hardware-level. Lower memory cost and faster matrix multiplication than those of continuous representations are attractive properties \cite{hubara2016binarized, rastegari2016xnor}.

However, because of the non-differentiable property of the discrete representation, a gradient could not be backward propagated through it. To overcome this challenge, the straight-through estimator (STE) was proposed \cite{bengio2013estimating}, which estimates the gradient for discretizing operations (e.g., sampling or \textit{argmax} operations) as 1 if the output is 1, otherwise it estimates 0. Although STE imposes a bias on the lower layer's gradient estimation, it is empirically demonstrated as an effective gradient estimator in the training of existing binary neural networks \cite{hubara2016binarized, rastegari2016xnor}. The STE estimator can be implemented with automatic differentiation tools such as Pytorch \cite{raiko2014techniques, Paszke2019pytorch}. With a smart reparameterization technique, it outputs a differentiable binary representation as follows.
\begin{align}
    s &\sim Bern(p), \quad s \in \{0, 1\}, \label{eq:binary_reparam_sampling} \\
    c &= s(1-p) + (1-s)(-p), \label{eq:binary_reparam_c} \\
    z &= p + \text{detach}(c), \label{eq:binary_reparam_z}
\end{align}
where $p$ is a normalized probability of the binary variable. $Bern(p)$ and $s$ are the Bernoulli distribution with a parameter $p$ and its random variable. $\text{detach}(c)$ is the operation that detaches its input, $c$, from the gradient computation graph in backward propagation stage. As a result, the final output, $z$, is a binary variable, $z \in \{0, 1\}$, and it can flow the gradient through the first term, $p$. As a variant of STE, it is easy to implement and work with other network architectures \cite{Rim2021arxiv}.

\section{Proposed Method: E2E BT}
In this section, we propose a new method, {\em E2E BT}, for end-to-end training of the BT framework like VAE. In Section~\ref{subsec:Categorical_Reparam_Trick}, we propose a new reparameterization trick to handle the non-differentiable property of the latent sentence. In Section~\ref{subsec:E2E_Backprop}, we derive objective functions for training in the BT framework with our proposed reparameterization trick. Also, we propose to use the language model's output distribution as an appropriate prior distribution of the latent sentence. Finally, in Section~\ref{subsec:regularization}, we propose a regularization technique that anneals the stochasticity of the latent sentence inference process during training.

\subsection{Categorical Reparameterization Trick (CRT)}
\label{subsec:Categorical_Reparam_Trick}
The distribution of a sentence is based on a sequence of categorical distributions of words, and a sampled sentence from the distribution is non-differentiable. To make backpropagation feasible through the sentence, we propose CRT, a reparameterization trick for categorical distribution, that is inspired by the binary reparameterization trick as in Section~\ref{subsec:Binary_Reparam_Trick}. Because we handle the categorical distribution (also called as Multinoulli distribution), the Bernoulli distribution in Eq.~\eqref{eq:binary_reparam_sampling} is replaced by the Multinoulli distribution with the probability $\pmb{p}$ computed by the inference model. Then, a one-hot vector for one word, $\pmb{s} \in \{0,1\}^{|V|}$ \textit{s.t.} $\sum_{i=1}^{|V|}{\pmb{s}_i}=1$, is sampled from the distribution, where $V$ is the vocabulary set. Instead of stochastic sampling based on a distribution, we can also use arbitrarily selection to determine $\pmb{s}$. It makes the CRT can output a one-hot vector for a class that does not have the highest probability. The CRT process is formulated as follow:
\begin{align}
    \pmb{s} &\sim Mult(\pmb{p}), \text{ or } \pmb{s} \text{ is given}, \label{eq:categ_reparam_sampling} \\
    \pmb{c} &= \pmb{s}\odot(\pmb{1}-\lambda\pmb{p}) + (\pmb{1}-\pmb{s})\odot(-\lambda\pmb{p}), \label{eq:categ_reparam_c} \\
    \pmb{z} &= \lambda\pmb{p} + \text{detach}(\pmb{c}), \label{eq:categ_reparam_z}
\end{align}
where $\odot$ is the element-wise multiplication. $Mult(\pmb{p})$ is the Multinoulli distribution given the normalized probability vector. Based on $\pmb{p}$, we compute the non-differentiable conjugate part, $\pmb{c}$, that is determined by the sample, $\pmb{s}$. 
Finally, it outputs a one-hot encoded vector $\pmb{z}$ which consists of $\pmb{p}$ and $\pmb{c}$ parts. $\pmb{c}$ is detached from the computation graph, so that backpropagation can flow gradient into the lower layers through $\pmb{p}$. We multiply the scalar $\lambda$ by $\pmb{p}$ in Eqs.\eqref{eq:categ_reparam_c} and \eqref{eq:categ_reparam_z} to control the amount of gradient that flows through $\pmb{p}$ while ensuring that the output $\pmb{z}$ remains a one-hot vector. Fig.~\ref{fig:illustration_our_method} illustrates the whole process of this trick in the BT framework. 

Given the Multinoulli distribution, $Mult(\pmb{p})$, the output process of the one-hot vector, $\pmb{s}$, is implemented by a sampling operator (e.g., Categorical sampling or argmax) and the one-hot encoding function that manually maps an integer value to a $|V|$-dimensional one-hot vector.

\subsubsection{Comparison between Gumbel-Softmax Trick (GST)}
\label{subsubsec:Comparison_between_Gumbel}
Like our proposed CRT, the GST is a reparameterization trick for categorical distribution \cite{jang2017categorical}. The GST process for $i$-th class is formulated as follow:
\begin{align}
    g_i &\sim Gumbel(0,1), \\
    z_i &= \frac{\exp{(\log{p_i}}+g_i)/\tau}{\sum_{j=1}^{|V|}\exp{(\log{p_j}}+g_j)/\tau}, \label{eq:gumbel_softmax}
\end{align}
where the scalar $\tau$ represents softmax temperature, which influences the sharpness of the output distribution in Eq.~\eqref{eq:gumbel_softmax}. Typically, a low value of $\tau$ is used. This choice encourages the distribution to approximate a one-hot vector by increasing the probability of the maximum value towards one, while decreasing the probabilities of other non-maximum values towards zero. In addition, the straight-through Gumbel-softmax trick (ST-GST) computes a strict one-hot vector while preserving the flow of gradient with using the trick below:
\begin{align}
    z_i^{st} &= \pmb{1}\left(\argmax_{j}z_j=i\right)-detach(z_i)+z_i. \label{eq:stgst}
\end{align}

Compared to the GST, we believe that the CRT has three benefits in the E2E BT training framework.
\begin{itemize}
\item \textbf{(1) Fast Computation}: the CRT has less computation cost than the GST (and ST-GST).
\item \textbf{(2) Controllable Gradient}: the CRT can control the amount of backward propagated gradient, while the GST cannot and the ST-GST needs modification.
\item \textbf{(3) Flexible Output}: the CRT is able to determine its one-hot vector output regardless of the distribution, while the GST (and ST-GST) cannot.
\end{itemize}
We will give explanations of the above benefits in detail.

First, the CRT is computationally less expensive because the GST operates the softmax function twice while the CRT operates it only once. The two softmax operations of the GST are as follows: one for computing the normalized word probability vector to match the scale with Gumbel distribution's sample and the other for the final output \cite{jang2017categorical}. The softmax function is expensive especially when the number of classes is large which is usually the case for natural language processing tasks. In experiments, we measured the spending times on reparameterization processes of the GST in Pytorch library's implementation \cite{Paszke2019pytorch} and our CRT. The vocabulary size, sentence length, and mini-batch size were set to 30000, 50, and 60, respectively. In the result, while the GST spent \textbf{3.35 seconds}, our CRT spent \textbf{0.98 seconds}, which is more than three times faster.

Also, our CRT can simply control the amount of backpropagated gradient. During the E2E BT training, the backpropagation might lead the inference model to learn undesirable degenerating solutions, such as copying for the easiest reconstruction. Therefore, the scale of the gradient needs to be adjusted to find a desirable solution. The CRT can adjust the backward propagated gradient with just multiplying the scalar $\lambda$ to the word probability in the reparameterization process, Eqs.\eqref{eq:categ_reparam_c} and \eqref{eq:categ_reparam_z}.
Then, $\lambda$ plays a role of the additional learning rate only for the inference model. We can set two coefficients, $\lambda_{x}$ and $\lambda_{y}$, for two languages, respectively, if they need different controls. In the GST, this trick is hard to implement, and the ST-GST needs the modification that multiplies $\lambda$ to the last two terms.

\begin{figure}[t!]
    \centering
    \includegraphics[width=0.6\linewidth]{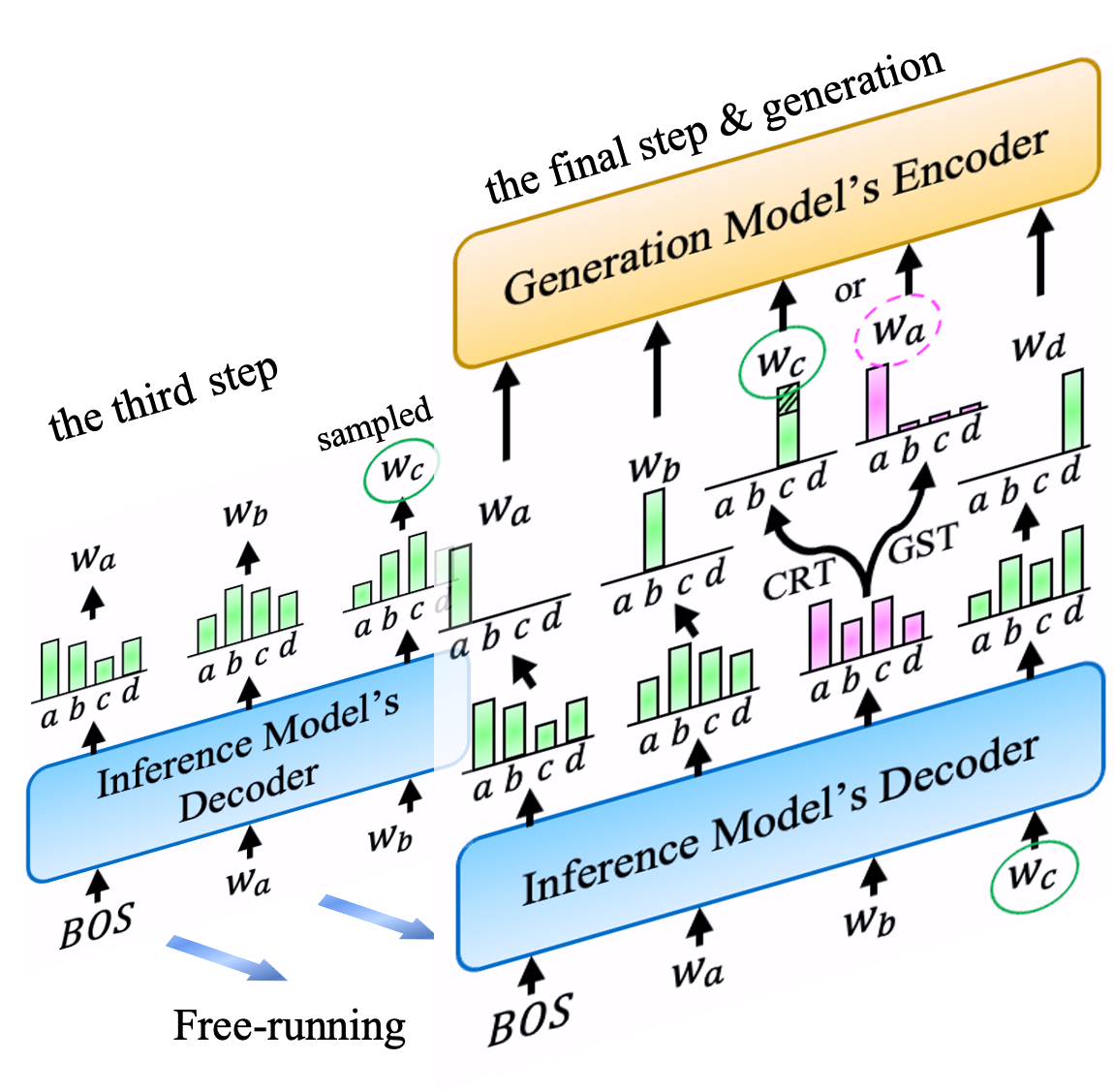}
    \caption{Example of the BT process with different reparameterization tricks, our CRT or the GST, where the third word's distribution at the third step (green) changed at the next step (pink). Our CRT can output the previous word, $w_c$, while the GST only outputs the maximally probable word, $w_a$, as the final output. 
    }
    \label{fig:illustration_discrepancy}
\end{figure}

Lastly, the CRT can reparameterize any word, while the GST and ST-GST reparameterize only the word that has the maximum probability. This is a crucial benefit, especially when the BT framework is implemented with the Transformer architecture \cite{vaswani2017attention} which is the standard model in various natural language processing tasks in these days. 
The inference process of the BT framework follows free-running where Transformer estimates every word's probability multiple times until the translation finishes. At every step, Transformer estimates the distributions of all words, including the previous output words. Thus, it is possible that the distribution of the previous word changes due to the stochasticity of the model such as dropout \cite{srivastava2014dropout}. That is, the maximally probable words in the final estimation might not be the same as the word sampled previously (See Fig. \ref{fig:illustration_discrepancy}). This situation may be even more frequent if we use stochastic sampling as the word sampling method. When there is discrepancy between the previous sampled words and the final words, it disturbs the gradient direction of the inference model in the wrong direction because the generation model computes and backpropagates the gradient based on the final output words, while the inference model receives the gradient assuming that it is computed based on the different words. 

Fig.~\ref{fig:illustration_discrepancy} gives an example with the vocabulary set, $\{w_a, w_b, w_c, w_d\}$. At the third step of the free-running inference, the word, $w_c$, is sampled from the third distribution (marked in green). However, in the next final step, the newly computed third distribution (marked in pink) is different from the previous one. Then, the GST samples a different word that has the maximum probability ($w_a$ in the figure) as the final output of the latent sentence, which is not the original conditional word, $w_c$, of the final estimation. Now, the generation model computes and backpropagates gradient through the last word, $w_d$, and it enforces the inference model to update its probability of $w_d$ given the final latent sentence, (BOS, $w_a$, $w_b$, $w_a$), as conditional words. However, the inference model will update its parameter assuming that the conditional words were (BOS, $w_a$, $w_b$, $w_c$). Contrary to the GST and ST-GST, our CRT can avoid this problem by just arbitrary selecting ('$\pmb{s}$ is given' in Eq.~\eqref{eq:categ_reparam_sampling}) $w_c$ instead of the maximum word.

We will analyze the feasibilities of the second and the last benefits in the experiment section to understand how those properties are crucial in the E2E BT training.

\subsection{End-to-End Training in the BT Framework}
\label{subsec:E2E_Backprop}
In this section, we describe our objective functions for our E2E BT training. The final objective function can be decomposed into two terms with two types of processes: bilingual and monolingual processes. The first objective term for the bilingual process is cross-entropy as follow:
\begin{align}
    \mathcal{J}_{ST}(\theta)&= -\sum_{(x^b,y^b) \in D^b} \log{p_{\theta}(y^b|x^b)}, \label{eq:Bilingual_J_ST}
\end{align}
where $x^b$ and $y^b$ are a source-language and its paired target-language sentences in the bilingual corpus, $D^b$, respectively. $\theta$ is the parameter of the ST-NMT model. Likewise, the objective function for the TS-NMT model, $\mathcal{J}_{TS}(\phi)$, is computed in the same way by switching the source and target sentences and replacing the $\theta$ with the TS-NMT model's parameter, $\phi$.

On the other hand, the objective function for the monolingual process is defined by the negative log likelihood of the monolingual sentence as follows:
\begin{align}
    \mathcal{J}_{TST} &= -\sum_{y^m \in D^y} \log{p(y^m)}, \label{eq:original_J_TST}
\end{align}
where $y^m$ is a target-language monolingual sentence in its monolingual corpus, $D^y$. The formulations are provided only for the TST process, but the STS process is simply symmetric to the TST process.

As usual, $\log{p(y^m)}$ in Eq. \ref{eq:original_J_TST} can be marginalized with latent sentences as follows:
\begin{align}
    \log{p(y^m)} &= \log{\sum_{\hat{x} \sim p(\hat{x})} p(y^m|\hat{x})p(\hat{x})}, \nonumber
\end{align}
where $\hat{x}$ is the inferred latent sentence that is the aligned representation of $y^m$ in the source-language domain, and $p(\hat{x})$ is a prior distribution of $\hat{x}$. By introducing an approximated posterior distribution, $q(\hat{x}|y)$, which is easy to sample $\hat{x}$ given $y^m$, we can derive the evidence lower bound objective (ELBO) of the marginal probability, $\log{p(y^m)}$, based on Jensen's inequality as follows:
\begin{align}
    \log{p(y^m)} &\geq \sum_{\hat{x} \sim q(\hat{x}|y^m)} q(\hat{x}|y^m)\log{ \frac{p(y^m|\hat{x})p(\hat{x})}{q(\hat{x}|y^m)}}, \nonumber \\
    \begin{split}
                &= \mathbb{E}_{\hat{x} \sim q(\hat{x}|y^m)}[\log{p(y^m|\hat{x})}] \\
                &\qquad- D_{KL}[q(\hat{x}|y^m) \parallel p(\hat{x})], \nonumber
    \end{split}
\end{align}
where $D_{KL}$ is Kullback-Liebler divergence (KL). In this TST process of the BT framework, the inference model, $q(\hat{x}|y^m)$, is modeled by the TS-NMT model with parameter $\phi$. Also, the generation model, $p(y^m|\hat{x})$, is modeled by the ST-NMT with parameter $\theta$.

As we mentioned in the introduction section \ref{sec:intro}, the latent sentence, $\hat{x}$, should be a valid sentence in the source language domain. Therefore, we use a language model's output distribution \cite{bengio2003neural,Paszke2019pytorch} as the prior distribution in the KL term, $p(\hat{x})$, instead of isotropic Gaussian in the conventional VAE. In practice, ELBO of the TST monolingual process can be formulated as follows:
\begin{align}
    \begin{split}
    \mathcal{J}^{ELBO}_{TST}(\phi,\theta) &= \sum_{y^m \in D^y} \mathbb{E}_{\hat{x} \sim q_{\phi}(\hat{x}|y^m)}[\log{p_{\theta}(y^m|\hat{x})}] \\
    &\quad- \alpha_{x}D_{KL}[q_{\phi}(\hat{x}|y^m) \parallel p_{\psi_{x}}(\hat{x})], \label{eq:ELBO_J_TST}
    \end{split}
\end{align}
where $p_{\psi_{x}}(\hat{x})$ is the pre-trained fixed source language model, and $\alpha_{x}$ is a hyperparameter to control the effect of the KL term in the final objective function.
Importantly, to make backpropagation feasible through the latent sentence, we apply the CRT to the latent sentence, $\hat{x}$. 
In order to input the latent sentence, which comprises a sequence of $|V|$-dimensional one-hot vectors, into the ST-NMT model, we implemented the conventional embedding layer as follows. While, in general, this layer accepts an integer value and returns the corresponding embedding vector from the embedding lookup table, we adapted the embedding layer to output a weighted sum of embedding vectors using the latent sentence vector's (one-hot) elements as the weight values.

Finally, the total objective functions for $\theta$ and $\phi$ are as follows:
\begin{align}
    \mathcal{J}_{T}(\theta, \phi) &= \mathcal{J}_{ST}(\theta) - \mathcal{J}^{ELBO}_{TST}(\phi, \theta),
    \\
    \mathcal{J}_{S}(\phi, \theta) &= \mathcal{J}_{TS}(\phi) - \mathcal{J}^{ELBO}_{STS}(\theta, \phi).
    \label{eq:total_obj_theta_phi}
\end{align}
The parameters of the both translation models are updated by the two objective functions.

\subsection{Regularization Technique: Annealing Stochasticity (AS)}
\label{subsec:regularization}
In this section, we propose an additional regularization technique related to the sampling method of the latent sentence. It would improve performance of our proposed original approach in Section~\ref{subsec:E2E_Backprop}. As argued in \cite{edunov2018understanding}, using stochastic sampling in the inference method gives more chances to find a better solution than the greedy method. However, adding more noise to the inference from the beginning of the training can cause the generation model to learn undesirable translation mapping, especially when the bilingual corpus is small. 
Considering the pro and con, we suggest to anneal the ratio of stochastic sampling during the training process as in the scheduled sampling approach in NMT \cite{bengio2015scheduled}. 
As an example of semi-online IBT framework, we newly infer a latent sentence given a monolingual sentence if we do not load the previous latent sentence from the memory. In that time, we infer the latent sentence by only the greedy method at the beginning and slowly increase the ratio of stochastic sampling as the training goes on.

\section{Experiments and Results}
Our experiments include internal ablation studies and comparisons with the previous BT works on two benchmark datasets: WMT18 English-German (En-De) and WMT18 English-Turkish (En-Tr) \cite{rej2018findings}, for large and small datasets, respectively. The bilingual corpora of `En-De' and `En-Tr' experiments consist of 5.2M and 0.2M sentence pairs, respectively. We collected 5.0M monolingual corpora for both languages of `En-De' and 4.7M monolingual corpora for `En-Tr' experiments. The whole monolingual corpora were randomly selected from NewsCrawl datasets provided by WMT18 translation task \cite{ott2018scaling}. About dataset pre-processing including tokenization, byte-pair encoding \cite{sennrich2015neural}, and making vocabularies, we follow the same processes with \cite{edunov2018understanding} based on the open source of fairseq toolkit\footnote{https://github.com/facebookresearch/fairseq/tree/main/examples/backtranslation}. We selected 32K and 10K most frequent subwords to make a vocabulary for each dataset, respectively. For validation of each experiment, we used Newstest13 (3K pairs) and Newsdev16 (1K pairs) for `En-De' and `En-Tr', respectively, from WMT18. Lastly, for testing, we used Newstest14$\sim$18 and Newstest16$\sim$18 for `En-De' and `En-Tr', respectively.

For the basic model architecture of the NMT models, we implemented the Transformer model with the base configuration of the original paper \cite{vaswani2017attention}. That is, the number of layers of encoder and decoder is 6. The dimensionality of the hidden state and word embedding dimension is 512. We refer the original paper for more specific configurations. As we mentioned in Section~\ref{subsec:E2E_Backprop}, we used the output distributions of language models as the prior distributions of the latent sentence of each language. We adopted two, English and German, Transformer-based language models \cite{Paszke2019pytorch}, which were pre-trained with the bilingual and monolingual corpora of `En-De' experiment. Likewise, other two, English and Turkish, Transformer-based language models were pre-trained with the bilingual and monolingual corpora of `En-Tr' experiment. We note that we did not use any additional data for the pre-training of language models.

For the optimization, we used the training strategy of fairseq toolkit \cite{ott2019fairseq}. The optimizer was Adam \cite{kingma2014adam} with the learning rate of 0.001. In addition, the inverse square root scheduler was used for the learning rate schedule. In each iteration, we made mini-batch consists of 8K and 4K tokens for `En-De' and `En-Tr' internal ablation study (Section~\ref{subsec:ablation_study}), respectively. For the comparisons with the previous BT works (Section~\ref{subsec:comparison_prev_bt}), we made mini-batch with 32K tokens for `En-De' experiments. We pre-trained the ST-NMT and TS-NMT with only bilingual corpus based on this optimization setting. Also, we used the same optimization setting for the pre-training of language models. 

For our main E2E BT training, we followed the semi-online IBT framework. Each mini-batch contains bilingual and monolingual tokens with 1:1 and 1:4 ratios for `En-De' and `En-Tr', respectively, within the total token sizes mentioned above. We set both $\lambda_x$ and $\lambda_y$ to 0.005 for both `En-De' and `En-Tr' experiments. We set 0.0005 value for both $\alpha_x$ and $\alpha_y$ of `En-De' experiment, and 0.0001 value for both $\alpha_x$ and $\alpha_y$ of `En-Tr' experiment. For the annealing schedule of AS regularization, we linearly increased the ratio of stochastic sampling from 0.0 to 1.0 during 300K iterations for the E2E BT training on the `En-De' dataset. Similarly, we set the ratio of stochastic sampling from 0.0 to 0.5 during 300K iterations for `En-Tr'.

\subsection{Internal Ablation Study}
\label{subsec:ablation_study}
In this section, we demonstrate the experimental results of our internal ablation study to understand the advantages of our proposed E2E BT training and the AS regularization on top of the bilingual NMT and the basic semi-online IBT (not E2E training) with the greedy selection method for latent sentence inference. In addition to the ablation study of `En-Tr' experiment, we also provide comparisons between our proposed CRT and the ST-GST as different reparameterization methods in E2E BT training. To evaluate translation quality, we used case-sensitive SacreBLEU \cite{post2018call} with `13a' tokenizer.

\begin{table}[]
\centering
\caption{Ablation study of `En-Tr' experiments in BLEU score. We used the beam search with a width of 5. In each evaluation, the left and right numbers of `/' mean En-to-Tr and Tr-to-En results. }
\resizebox{0.85\textwidth}{!}{
\begin{tabular}{|c|c|c|c|c|}
\hline
Model       & Newstest16  & Newstest17  & Newstest18  & Average \\ 
\hhline{|=====|}
Transformer & 11.54/15.62 & 12.11/15.66 & 10.58/16.57 & 11.41/15.95 \\ 
\hline
BasicBT     & 16.26/22.67 & 18.69/22.10 & 15.37/23.15 & 16.77/22.64 \\ 
\hline
BasicBT+AS  & 16.50/22.48 & 18.74/21.86 & 15.58/23.35 & 16.94/22.56 \\ 
\hline
E2E BT+AS   & \textbf{16.68/23.44} & \textbf{19.29/22.28} & \textbf{15.76/24.12} & \textbf{17.24/23.28} \\ 
\hline
\end{tabular}}
\label{table:entr_ablation_study_results}
\end{table}

Table~\ref{table:entr_ablation_study_results} presents the BLEU score results on each testset of each model. `Transformer' indicates the NMT model that was trained with only bilingual corpus. `BasicBT' is the basic semi-online IBT method that does not train the two NMT models (ST-NMT and TS-NMT) in an end-to-end fashion. On top of `BasicBT', we applied AS first, and then E2E BT, because the original VAE uses stochastic sampling to infer the latent variable during training, but `BasicBT' is based on greedy selection. Therefore, before applying E2E BT, we applied AS to give appropriate amount of stochasticity during inference at each iteration. Our proposed AS regularization technique shows a slight improvement in En-to-Tr translation. Interestingly, when we apply E2E BT training to the `BasicBT+AS' model, it noticeably increased BLEU scores in general.

\begin{table}[]
\centering
\caption{Comparisons of reparameterization methods in BLEU score. We used the beam search with a width of 5. In each evaluation, the left and right numbers of `/' mean En-to-Tr and Tr-to-En results. `E2E BT w/ CRT' is the same model with the `E2E BT+AS' in Table~\ref{table:entr_ablation_study_results}}
\resizebox{0.9\textwidth}{!}{
\begin{tabular}{|l|c|c|c|c|}
\hline
\multicolumn{1}{|c|}{Model} & Newstest16 & Newstest17 & Newstest18 & Average \\ 
\hhline{|=====|}
E2E BT w/ CRT              & 16.68/23.44 & 19.29/22.28 & 15.76/24.12 & 17.24/23.28 \\ 
\hline
E2E BT w/ ST-GST           & 2.51/3.61   & 2.54/3.02   & 2.42/3.99   & 2.49/3.54   \\ 
\hline
E2E BT w/ ST-GST+$\lambda$ & 15.66/22.63 & 17.84/21.28 & 14.92/23.73 & 16.15/22.55 \\ 
\hline
E2E BT w/ ST-GST+$\lambda$+FO   & 16.48/23.26 & 19.03/22.14 & 15.83/24.24 & 17.11/23.21 \\ 
\hline
\end{tabular}}
\label{table:entr_stgst_comparison}
\end{table}

Table~\ref{table:entr_stgst_comparison} demonstrates the comparison between several reparameterization methods. First, we find that the ST-GST reparameterization method, `E2E BT w/ ST-GST', is inappropriate in this E2E BT training, leading to significantly low BLEU scores. We interpret the absences of the last two advantageous properties of the CRT, namely `(2) Controllable Gradient' and `(3) Flexible Output (FO)' which were discussed in Section~\ref{subsubsec:Comparison_between_Gumbel}, as significant factors contributing to the poor performances. 
To check the feasibilities of these two properties, we modified ST-GST to acquire those properties. First, as in CRT, we multiplied $\lambda$ to the last two terms of Eq.~\eqref{eq:stgst}, `E2E BT w/ ST-GST+$\lambda$', whose $\lambda$ is set to 0.0005 as in `E2E BT w/ CRT'. We find that the gradient controllable property is the most crucial factor in the E2E BT training, because it can prevent the model to learn the degenerating solution (refer Section~\ref{subsubsec:Comparison_between_Gumbel}). However, the performances of `E2E BT w/ ST-GST+$\lambda$' are still lower than `E2E BT w/ CRT'. Therefore, we further modified the ST-GST to ensuring the FO property with adding the value instead of $g_i$ in Eq.~\eqref{eq:gumbel_softmax} that ensures the arbitrarily selected word has the highest value regardless of the output probability. We modeled the value as follow:
\begin{align}
    b_i =&
        \begin{cases}
            \log{\frac{\max\{p_1,\dots,p_{|V|}\}}{\min\{p_1,\dots,p_{|V|}\}}}+\epsilon & \text{if } i \text{ is the index to reparam.}, \\
            0 & \text{otherwise}, \\
        \end{cases}
\end{align}
where $\epsilon$ is a small value of scalar. Finally, when we applied the final modified version of ST-GST to E2E BT, `E2E BT w/ ST-GST+$\lambda$+FO', it achieved similar performances with `E2E BT w/ CRT'. We believe that the current ST-GST needs such modifications to work in E2E BT training, while our proposed CRT is faster and more appropriate to the E2E BT training.

\begin{table}[]
\caption{Ablation study of `En-De' experiments in BLEU score. We used the beam search with a width of 5. In each evaluation, the left and right numbers of `/' mean En-to-De and De-to-En results. }
\centering
\resizebox{1.0\textwidth}{!}{
\begin{tabular}{|c|c|c|c|c|c|c|}
\hline
Model       & \multicolumn{1}{c|}{Newstest14} & \multicolumn{1}{c|}{Newstest15} & \multicolumn{1}{c|}{Newstest16} & \multicolumn{1}{c|}{Newstest17} & \multicolumn{1}{c|}{Newstest18} & \multicolumn{1}{c|}{Average} \\
\hhline{|=======|}
Transformer & 24.74/29.65 & 28.24/30.76 & 32.64/36.06 & 26.45/31.47 & 39.32/38.38 & 30.28/33.26 \\ \hline
BasicBT     & 27.87/30.01 & 29.80/31.99 & 34.42/39.44 & 28.33/33.11 & 41.26/41.08 & 32.37/35.13 \\ \hline
Basic BT+AS & \textbf{28.27}/32.21 & 30.26/33.48 & 34.58/40.88 & 28.48/\textbf{34.98} & 41.71/42.47 & 32.66/36.80 \\ \hline
E2E BT+AS   & 28.26/\textbf{32.63} & \textbf{30.54}/\textbf{33.98} & \textbf{34.70}/\textbf{41.07} & \textbf{28.64}/34.89 & \textbf{42.15}/\textbf{42.51} & \textbf{32.86}/\textbf{37.02} \\ \hline
\end{tabular}}
\label{table:ende_ablation_study_results}
\end{table}

Table~\ref{table:ende_ablation_study_results} presents the ablation study of `En-De' experiments. Macroscopically, our proposed AS regularization and E2E BT training shows similar tendencies of performance gains with `En-Tr' experiments. However, AS regularization, in this experiment, shows noteworthy performance gain in `De-En' more than `En-De'. Based on these experiments, we argue that our proposed approaches are advantageous in several benchmark datasets.

\subsection{Comparisons with Previous BT Works}
\label{subsec:comparison_prev_bt}
In this section, we compare our approaches with other recent BT works that are based on base Transformer architectures \cite{zheng2019mirror,gracca2019generalizing,xu2020dual,pham2021meta}. However, there are many differences on testing such as different testsets, different BLEU score metrics, etc. For fair comparisons, we report the average BLEU scores of the previous work and our method on their testsets with the same BLEU metric. Also, we report the size of monolingual corpus that they used for BT training (note that we used 5M monolingual corpora for `En-De' experiments).

\begin{table}[]
\caption{Comparisons with the previous BT works that conducted `En-De' experiments based on the base Transformer architecture. In each evaluation, the left and right numbers of `/' mean En-to-De and De-to-En averaged BLEU scores. Based on the open source code, we found the BLEU score metric that (Z. Zheng et al. 2019) used was SacreBLEU with `intl' tokenizer and case-insensitive. We could not find the specific information of (H. Pham et al. 2021)'s BLEU score metric, so we used tokenBLEU for the comparison.}
\centering
\resizebox{1.0\textwidth}{!}{
\begin{tabular}{|c|c|c|c|l|l|}
\hline
Previous Work          & Newstest   & \begin{tabular}[c]{@{}c@{}}BLEU\\ Metric\end{tabular} & \begin{tabular}[c]{@{}c@{}}\# of \\ MonoData\end{tabular} & \multicolumn{1}{c|}{\begin{tabular}[c]{@{}c@{}}Avg. BLEU\\ (Prev.Work)\end{tabular}} & \multicolumn{1}{c|}{\begin{tabular}[c]{@{}c@{}}Avg. BLEU\\ (Our Work)\end{tabular}} \\ 
\hhline{|======|}
(Z. Zheng et al. 2019)\cite{zheng2019mirror} & 14 & SacreBLEU & 5M & 30.30/33.80 & 30.37/34.21 \\ \hline
(M. Gra\c{c}a et al. 2019)\cite{gracca2019generalizing} & 17 & SacreBLEU & 4M & 28.60/ -  & 29.17/35.58  \\ \hline
(W. Xu et al. 2020)\cite{xu2020dual}    & 16$\sim$18 & SacreBLEU & 5M & 30.00/33.05 & 35.54/39.79 \\ \hline
(H. Pham et al. 2021)\cite{pham2021meta}  & 14 & unknown & 220M & 30.39/ - & 30.77/33.97 \\ \hline
\end{tabular}}
\label{table:comparison_prev_bt}
\end{table}

Table~\ref{table:comparison_prev_bt} demonstrates the comparisons between a previous work's average BLEU score and the same one of our proposed E2E BT model with AS regularization based on the previous work's BLEU score metric and testsets. We found that our work steadily outperforms the previous works. Especially, the comparison with (H. Pham et al. 2021) indicates an astonishing efficiency of our proposed approach when we consider the amount of monolingual data they used.

\section{Conclusion}
In this paper, we proposed a categorical reparameterization trick that makes the translated sentences differentiable. Based on the trick, backpropagation became feasible through the sentences, and we used this trick to train both translation models in the end-to-end learning fashion for back-translation. To train the models together in back-translation, we developed the evidence lower bound objective, which could train the translation models in the semi-supervised learning fashion.  
In addition, we proposed a regularization techniques that are practically advantageous in the back-translation training. Finally, our experimental results demonstrated that our proposed method is beneficial to learn better translation models while outperforming the baselines. 


\subsubsection{Acknowledgements} 
This research was supported by Basic Science Research Program through the National Research Foundation of Korea funded by the Ministry of Education (NRF-2022R1A2C1012633)

\bibliography{icann24}

\begin{thebibliography}{10}
\providecommand{\url}[1]{\texttt{#1}}
\providecommand{\urlprefix}{URL }
\providecommand{\doi}[1]{https://doi.org/#1}

\bibitem{bahdanau2014neural}
Bahdanau, D., Cho, K., Bengio, Y.: Neural machine translation by jointly learning to align and translate. arXiv preprint arXiv:1409.0473  (2014)

\bibitem{bengio2015scheduled}
Bengio, S., Vinyals, O., Jaitly, N., Shazeer, N.: Scheduled sampling for sequence prediction with recurrent neural networks. Advances in neural information processing systems  \textbf{28} (2015)

\bibitem{bengio2003neural}
Bengio, Y., Ducharme, R., Vincent, P., Jauvin, C.: A neural probabilistic language model. Journal of machine learning research  \textbf{3}(Feb),  1137--1155 (2003)

\bibitem{bengio2013estimating}
Bengio, Y., L{\'e}onard, N., Courville, A.: Estimating or propagating gradients through stochastic neurons for conditional computation. arXiv preprint arXiv:1308.3432  (2013)

\bibitem{rej2018findings}
Bojar, O., Federmann, C., Fishel, M., Graham, Y., Haddow, B., Koehn, P., Monz, C.: Findings of the 2018 conference on machine translation (wmt18). In: Proceedings of the Third Conference on Machine Translation. vol.~2, pp. 272--307 (2018)

\bibitem{caswell2019tagged}
Caswell, I., Chelba, C., Grangier, D.: Tagged back-translation. arXiv preprint arXiv:1906.06442  (2019)

\bibitem{cotterell2018explaining}
Cotterell, R., Kreutzer, J.: Explaining and generalizing back-translation through wake-sleep. arXiv preprint arXiv:1806.04402  (2018)

\bibitem{currey2017copied}
Currey, A., Miceli-Barone, A.V., Heafield, K.: Copied monolingual data improves low-resource neural machine translation. In: Proceedings of the Second Conference on Machine Translation. pp. 148--156 (2017)

\bibitem{domhan2017using}
Domhan, T., Hieber, F.: Using target-side monolingual data for neural machine translation through multi-task learning. In: Proceedings of the 2017 Conference on Empirical Methods in Natural Language Processing. pp. 1500--1505 (2017)

\bibitem{edunov2018understanding}
Edunov, S., Ott, M., Auli, M., Grangier, D.: Understanding back-translation at scale. arXiv preprint arXiv:1808.09381  (2018)

\bibitem{fadaee2018back}
Fadaee, M., Monz, C.: Back-translation sampling by targeting difficult words in neural machine translation. arXiv preprint arXiv:1808.09006  (2018)

\bibitem{Goodfellow-et-al-2016}
Goodfellow, I., Bengio, Y., Courville, A.: Deep Learning. MIT Press (2016), \url{http://www.deeplearningbook.org}

\bibitem{gracca2019generalizing}
Gra{\c{c}}a, M., Kim, Y., Schamper, J., Khadivi, S., Ney, H.: Generalizing back-translation in neural machine translation. arXiv preprint arXiv:1906.07286  (2019)

\bibitem{gulcehre2015using}
Gulcehre, C., Firat, O., Xu, K., Cho, K., Barrault, L., Lin, H.C., Bougares, F., Schwenk, H., Bengio, Y.: On using monolingual corpora in neural machine translation. arXiv preprint arXiv:1503.03535  (2015)

\bibitem{guo2021revisiting}
Guo, Y., Zhu, H., Lin, Z., Chen, B., Lou, J.G., Zhang, D.: Revisiting iterative back-translation from the perspective of compositional generalization. In: Proceedings of the AAAI Conference on Artificial Intelligence. vol.~35, pp. 7601--7609 (2021)

\bibitem{hahn2019disentangling}
Hahn, S., Choi, H.: Disentangling latent factors of variational auto-encoder with whitening. In: International Conference on Artificial Neural Networks. pp. 590--603. Springer (2019)

\bibitem{han2021unsupervised}
Han, J.M., Babuschkin, I., Edwards, H., Neelakantan, A., Xu, T., Polu, S., Ray, A., Shyam, P., Ramesh, A., Radford, A., et~al.: Unsupervised neural machine translation with generative language models only. arXiv preprint arXiv:2110.05448  (2021)

\bibitem{he2016dual}
He, D., Xia, Y., Qin, T., Wang, L., Yu, N., Liu, T.Y., Ma, W.Y.: Dual learning for machine translation. Advances in neural information processing systems  \textbf{29},  820--828 (2016)

\bibitem{Higgins2017betaVAELB}
Higgins, I., Matthey, L., Pal, A., Burgess, C.P., Glorot, X., Botvinick, M.M., Mohamed, S., Lerchner, A.: Beta-vae: Learning basic visual concepts with a constrained variational framework. In: ICLR (2017)

\bibitem{hoang2018iterative}
Hoang, V.C.D., Koehn, P., Haffari, G., Cohn, T.: Iterative back-translation for neural machine translation. In: Proceedings of the 2nd Workshop on Neural Machine Translation and Generation. pp. 18--24 (2018)

\bibitem{hubara2016binarized}
Hubara, I., Courbariaux, M., Soudry, D., El-Yaniv, R., Bengio, Y.: Binarized neural networks. Advances in neural information processing systems  \textbf{29} (2016)

\bibitem{imamura2018enhancement}
Imamura, K., Fujita, A., Sumita, E.: Enhancement of encoder and attention using target monolingual corpora in neural machine translation. In: Proceedings of the 2nd Workshop on Neural Machine Translation and Generation. pp. 55--63 (2018)

\bibitem{jang2017categorical}
Jang, E., Gu, S., Poole, B.: Categorical reparametrization with gumble-softmax. In: International Conference on Learning Representations (ICLR 2017). OpenReview. net (2017)

\bibitem{kim2018disentangling}
Kim, H., Mnih, A.: Disentangling by factorising. In: International Conference on Machine Learning. pp. 2649--2658. PMLR (2018)

\bibitem{kingma2014adam}
Kingma, D.P., Ba, J.: Adam: A method for stochastic optimization. arXiv preprint arXiv:1412.6980  (2014)

\bibitem{kingma2013auto}
Kingma, D.P., Welling, M.: Auto-encoding variational bayes. arXiv preprint arXiv:1312.6114  (2013)

\bibitem{ott2019fairseq}
Ott, M., Edunov, S., Baevski, A., Fan, A., Gross, S., Ng, N., Grangier, D., Auli, M.: fairseq: A fast, extensible toolkit for sequence modeling. In: Proceedings of NAACL-HLT 2019: Demonstrations (2019)

\bibitem{ott2018scaling}
Ott, M., Edunov, S., Grangier, D., Auli, M.: Scaling neural machine translation (2018)

\bibitem{Paszke2019pytorch}
Paszke, A., Gross, S., Massa, F., Lerer, A., Bradbury, J., Chanan, G., Killeen, T., Lin, Z., Gimelshein, N., Antiga, L., Desmaison, A., Kopf, A., Yang, E., DeVito, Z., Raison, M., Tejani, A., Chilamkurthy, S., Steiner, B., Fang, L., Bai, J., Chintala, S.: Pytorch: An imperative style, high-performance deep learning library. In: Wallach, H., Larochelle, H., Beygelzimer, A., d\textquotesingle Alch\'{e}-Buc, F., Fox, E., Garnett, R. (eds.) Advances in Neural Information Processing Systems 32, pp. 8024--8035. Curran Associates, Inc. (2019)

\bibitem{pham2021meta}
Pham, H., Wang, X., Yang, Y., Neubig, G.: Meta back-translation. arXiv preprint arXiv:2102.07847  (2021)

\bibitem{post2018call}
Post, M.: A call for clarity in reporting {BLEU} scores. In: Proceedings of the Third Conference on Machine Translation: Research Papers. pp. 186--191. Association for Computational Linguistics, Belgium, Brussels (Oct 2018), \url{https://www.aclweb.org/anthology/W18-6319}

\bibitem{raiko2014techniques}
Raiko, T., Berglund, M., Alain, G., Dinh, L.: Techniques for learning binary stochastic feedforward neural networks. arXiv preprint arXiv:1406.2989  (2014)

\bibitem{rastegari2016xnor}
Rastegari, M., Ordonez, V., Redmon, J., Farhadi, A.: Xnor-net: Imagenet classification using binary convolutional neural networks. In: European conference on computer vision. pp. 525--542. Springer (2016)

\bibitem{Rim2021arxiv}
Rim, D.N., Jang, I., Choi, H.: Deep neural networks and end-to-end learning for audio compression. arXiv preprint arXiv:2105.11681  (2021)

\bibitem{sennrich2015improving}
Sennrich, R., Haddow, B., Birch, A.: Improving neural machine translation models with monolingual data. arXiv preprint arXiv:1511.06709  (2015)

\bibitem{sennrich2015neural}
Sennrich, R., Haddow, B., Birch, A.: Neural machine translation of rare words with subword units. arXiv preprint arXiv:1508.07909  (2015)

\bibitem{skorokhodov2018semi}
Skorokhodov, I., Rykachevskiy, A., Emelyanenko, D., Slotin, S., Ponkratov, A.: Semi-supervised neural machine translation with language models. In: Proceedings of the AMTA 2018 workshop on technologies for MT of low resource languages (LoResMT 2018). pp. 37--44 (2018)

\bibitem{srivastava2014dropout}
Srivastava, N., Hinton, G., Krizhevsky, A., Sutskever, I., Salakhutdinov, R.: Dropout: a simple way to prevent neural networks from overfitting. The journal of machine learning research  \textbf{15}(1),  1929--1958 (2014)

\bibitem{vaswani2017attention}
Vaswani, A., Shazeer, N., Parmar, N., Uszkoreit, J., Jones, L., Gomez, A.N., Kaiser, {\L}., Polosukhin, I.: Attention is all you need. In: Advances in neural information processing systems. pp. 5998--6008 (2017)

\bibitem{xu2020dual}
Xu, W., Niu, X., Carpuat, M.: Dual reconstruction: a unifying objective for semi-supervised neural machine translation. arXiv preprint arXiv:2010.03412  (2020)

\bibitem{zhang2016exploiting}
Zhang, J., Zong, C.: Exploiting source-side monolingual data in neural machine translation. In: Proceedings of the 2016 Conference on Empirical Methods in Natural Language Processing. pp. 1535--1545 (2016)

\bibitem{zhang2018joint}
Zhang, Z., Liu, S., Li, M., Zhou, M., Chen, E.: Joint training for neural machine translation models with monolingual data. In: Thirty-Second AAAI Conference on Artificial Intelligence (2018)

\bibitem{zheng2019mirror}
Zheng, Z., Zhou, H., Huang, S., Li, L., Dai, X.Y., Chen, J.: Mirror-generative neural machine translation. In: International Conference on Learning Representations (2019)

\end{thebibliography}
\bibliographystyle{splncs04}

\end{document}